# Rethinking Cognition: Morphological Info-Computation and the Embodied Paradigm in Life and Artificial Intelligence

Gordana Dodig-Crnkovic[1,2]

[1] Department of Computer Science and Engineering, Chalmers University of Technology, Gothenburg, Sweden.
dodig@chalmers.se
[2] Division of Computer Science and Software Engineering, School of Innovation, Design and Engineering, Mälardalen University, Västerås, Sweden

**Abstract**. This study aims to place Lorenzo Magnani's Eco-Cognitive Computationalism within the broader context of current work on information, computation, and cognition. Traditionally, cognition was believed to be exclusive to humans and a result of brain activity. However, recent studies reveal it as a fundamental characteristic of all life forms, ranging from single cells to complex multicellular organisms and their networks. Yet, the literature and general understanding of cognition still largely remain human-brain-focused, leading to conceptual gaps and incoherency. This paper presents a variety of computational (information processing) approaches, including an info-computational approach to cognition, where natural structures represent information and dynamical processes on natural structures are regarded as computation, relative to an observing cognizing agent. We model cognition as a web of concurrent morphological computations, driven by processes of self-assembly, self-organization, and autopoiesis across physical, chemical, and biological domains. We examine recent findings linking morphological computation, morphogenesis, agency, basal cognition, extended evolutionary synthesis, and active inference. We establish a connection to Magnani's Eco-Cognitive Computationalism and the idea of computational domestication of ignorant entities. Novel theoretical and applied insights question the boundaries of conventional computational models of cognition. The traditional models prioritize symbolic processing and often neglect the inherent constraints and potentialities in the physical embodiment of agents on different levels of organization. Gaining a better info-computational grasp of cognitive embodiment is crucial for the advancement of fields such as biology, evolutionary studies, artificial intelligence, robotics, medicine, and more.

## Lorenzo Magnani's Eco-cognitive computationalism and the idea of computational domestication of ignorant entities

Lorenzo Magnani's Eco-Cognitive Computationalism (Magnani, 2018) (Magnani, 2021) (Magnani, 2022b) (Arfini & Magnani, 2022) is a philosophical framework that explores the interplay between cognitive processes and their environmental contexts, particularly emphasizing the role of computational tools in enhancing human cognitive abilities. The concept of "computational domestication of ignorant entities" within this framework can be interpreted as the use of computational methods and models to harness and transform less understood or unorganized aspects of the environment (which he refers to as "ignorant entities") into structured, intelligible forms that can be more easily manipulated and understood by human cognition.

This approach suggests that computational tools not only extend our cognitive capabilities but also reshape our cognitive environment, making it more open to human understanding and control. By "domesticating" these ignorant entities through computational means, we effectively restructure our cognitive landscape, enhancing our ability to solve complex problems and generate new knowledge.

Magnani's theory touches on themes common in the philosophy of cognitive science, such as the *extended mind thesis* and *situated cognition*, but places a unique emphasis on the *transformative power of computational technology* in these processes. This perspective is particularly relevant in the context of today's data-driven science and technology, where computational tools play a critical role in organizing, analyzing, and interpreting large datasets, effectively bringing clarity and insights from what might initially appear as chaotic or overwhelming information.

In the contemporary landscape of cognitive science and artificial intelligence, Lorenzo Magnani's book "Eco-Cognitive Computationalism: Cognitive Domestication of Ignorant Entities" presents a groundbreaking perspective. The concept of eco-cognitive computationalism offers a novel approach to understanding computation as an active process in physical entities, which are *transformed into cognitive mediators*. The main themes and concepts presented in Magnani's work emphasize the dynamic interplay between internal processes, the body, and the environment in the emergence of information, computation, and cognition. Eco-cognitive computationalism views computation as an active process in distributed physical entities that are transformed to encode and decode data. This approach *moves* beyond the traditional notion of digital computation, encompassing a broader range of computational activities performed by physical devices, including the brain. Magnani argues that this perspective avoids reducing the analysis of computation to a rigid, one-sided view, instead proposing an intellectual framework that respects the historical and dynamic nature of concepts of information, computation, and cognition (Magnani, 2022b).

One of the central themes in Magnani's approach is the evolutionary emergence of information and cognition as the result of dynamic interactions *between internal processes, the body, and the external environment*. Magnani emphasizes that understanding computation requires acknowledging its context within the ecology of cognition, which includes both organic and artefactual agents (Magnani, 2022b). Turing's insights into the emergence of cognition and computation are fundamental in this discussion. Turing's concept of *unorganized machines, which can be educated* through interaction with their environment, illustrates the dynamic nature of cognitive development. This analogy between human brains and computational machines underscores the *coevolution of information, computation, and cognition* in both natural and artificial systems (Turing, 1948).

Magnani extends the discussion of eco-cognitive computationalism to include embodied and distributed cognition. He argues that cognitive processes are not confined to the brain but are distributed across the body and external environment. This view aligns with recent cognitive science theories that emphasize the role of the body and environment in shaping cognitive tasks, (Clark, 1989, 1997, 2008, 2016). The creation of *mimetic bodies*, which enhance computational capabilities through *morphological features*, exemplifies this approach. These bodies simplify cognitive and motor tasks, highlighting the *interdependence of internal and external resources* in cognitive processes.

Magnani (Magnani, 2022b) and (Magnani, 2022a) details the cognitive strategies employed by deep learning machines, such as AlphaGo. Magnani distinguishes between locked and unlocked strategies, examining how these approaches impact the creativity and adaptability of AI systems. Locked strategies, which are limited in their eco-cognitive openness, contrast with the more flexible and creative reasoning observed in human cognition.

The final chapter in (Magnani, 2022b) explores the *domestication* of physical and biological bodies through computational means. Magnani introduces the concept of *overcomputationalism*, which critiques the tendency to overextend computational applications to all aspects of life. He emphasizes the need to *protect and preserve the unique qualities of ignorant entities*, advocating for a balance between computational and human-centered approaches. The future of eco-cognitive settings, Magnani argues, should be tailored to enhance both computational efficiency and human well-being.

Magnani's Eco-cognitive computationalism presents a comprehensive and dynamic framework for understanding computation, information, and cognition. By integrating insights from cognitive science, artificial intelligence, and philosophy, Magnani presents a holistic view of computational systems as integral to both natural and artificial environments. His emphasis on the dynamic and distributed nature of cognition and computation challenges traditional notions and opens new perspectives for research and application in these fields.

# Computation in a computing Universe. Nature as a network of morphological info-computational processes (ICON) for cognitive agents

Magnani's Eco-cognitive computationalism with cognitive domestication of ignorant entities can be related to a broader computational perspective of Info-computationalism, which addresses nature in computational terms. The info-computational approach (ICON) presents a view of nature as a network of information-processing agents organized in a dynamic hierarchy of levels. This framework aims to unify diverse understandings of natural, formal, technical, behavioral, and social phenomena through the perspective of information and computation. The idea that, for us, information is the fabric of reality goes back to Wheeler's "it from bit" (Wheeler, 1990). Our reality is a result of the processing of information of multimodal signals coming through our senses combined with information processes in our bodies that unfold on several levels of organization/scale. Among prominent works on informational reality are (Baeyer von, 2004)(Lloyd, 2006)(Seife, 2006)(Vedral, 2010)(Davies & Gregersen, 2010)(Burgin, 2010). A world exists on its own with human agents in it and we discover its structures and processes, patterns and laws, by interactions and via self-structuring (self-organization) of information obtained through morphological/physical computation processes (Pfeifer & Iida, 2005) (Pfeifer & Gomez, 2009) (Dodig-Crnkovic, 2013). In this analysis, a natural philosophy framework is presented that provides a foundation for connecting domain-specific knowledge obtained from a variety of sciences and other knowledge fields into an interdisciplinary/transdisciplinary perspective.

Unlike old-day stand-alone calculating machines, represented by the abstract logical Turing Machine Model, modern networked and concurrently executing computers are used for worldwide communication, information processing, cyber-physical control, and data-, information-, and knowledge management. They serve as cognitive tools of the extended mind (Clark & Chalmers, 1998a), facilitating social interactions, decision-making, and control of physical processes. This interconnected concurrent nature of computation aligns with the actor model of computation (Hewitt, 2007, 2012), emphasizing distributed, reactive, agent-based, and parallel execution. Such systems operate continuously, focusing on behavior, response to changes, and adaptability rather than predetermined algorithm termination.

*History of the idea of the computing universe*

Konrad Zuse first suggested that the physical behavior of the universe could be computed on a basic level, using cellular automata (Zuse, 1970). A similar, pan-computationalist view is supported by various scientists (Wheeler, 1990) (Wheeler, 1994)(Fredkin, 1990), (Wolfram, 2002) (Dodig-Crnkovic, 2012) who see natural phenomena as results of computational processes. This perspective aligns with the idea that the universe computes its next state from its current state, with interactions and information exchanges driving its evolution (Chaitin, 2006).

*Natural, physical, and morphological computation*

Natural computation encompasses a range of computational systems inspired by nature, including neural networks, genetic algorithms, and quantum computing. This field promoted a generalized model of computation beyond traditional Turing machines, recognizing the complexity of biological and cognitive systems. Morphological computation, a key concept, explains how physical interactions lead to the development of complex structures and processes in living systems.
(Dodig-Crnkovic, 2013)(Dodig-Crnkovic, 2017); (Dodig-Crnkovic & Lowe, 2017) (Pfeifer et al., 2006) (Hauser et al., 2014) (Ghazi-Zahedi et al., 2017) (Pfeifer & Iida, 2005).

Natural info-computation provides a unified framework for understanding cognitive systems through morphological info-computation across various levels of organization. It advocates for a new philosophy of nature, integrating information and computation to explain diverse phenomena. By generalizing computational models, this approach aims to encompass the full range of natural processes, from physics to cognition, highlighting the essential role of agents in these systems. The framework necessitates generalizing computational models beyond the traditional Turing machine model, requiring agent-based concurrent resource-sensitive models to cover phenomena from physics to cognition.

In this approach, nature computes on a variety of levels of organization where informational structures (as observed by diverse cognizing agents at different levels) dynamically change through interactive processes of natural computation. Natural computing (Rozenberg & Kari, 2008) (Rozenberg et al., 2012) encompasses a wide range of areas, including quantum computing, chemical computing, biological computing, evolutionary algorithms, swarm intelligence, reservoir computing, thermodynamic computing, and cognitive and intelligent computing. Each of these areas draws on principles and phenomena observed in the natural world to inform and improve computational techniques.

Natural computing (unconventional computing) raises philosophical questions about the nature of computation and intelligence, as it challenges traditional notions about where and how computation can occur. Natural computing continues to grow as a field of research that bridges the gap between natural sciences and computer science, enabling new solutions to complex problems by using the power of natural processes. It exemplifies a profound shift toward computing that is not confined to silicon but is increasingly interwoven with biological and quantum systems, (Denning, 2003, 2007, 2009, 2010) (Paun et al., 2017; Rozenberg, 2008; Rozenberg & Kari, 2008).

*The necessity of an agent/observer/actor for the generation of knowledge*

Information, a concept widely used and ambiguous, is essential for understanding both everyday applications and formal definitions in various fields of research. Current understanding of matter/energy also reflects ambiguity, as observable matter constitutes only a small percentage of the universe, with the rest attributed to dark matter and dark energy. This indicates a need to re-examine our understanding of the physical world. What is missing in the traditional picture is the model of a physical observer, such as proposed in second-order cybernetics. As (Goyal, 2012) explains, an observer in classical physics is a highly idealized omnipotent agent looking at the world from nowhere (a "God's eyes" perspective). Or, as (Fields, 2012) expresses it, a "Galilean observer" is a bare "point of view", able to observe and conceptualize the world without affecting it or being affected. There is a hope to gain more clarity on the structures and dynamics of the universe by connecting information and matter/energy. Information is defined as differences that make a difference, (Bateson, 1972), or "Information expresses the fact that a system is in a certain configuration that is correlated to the configuration of another system. Any physical system may contain information about another physical system." (Hewitt, 2007) In other words, information is *relational*, which is fundamental.

*Relational info-computation as a reality for an agent*

> "What we call reality arises in the last analysis from the posing of yes-no questions and the registering of equipment-evoked responses; in short, all things physical are information-theoretic in origin and this is a participatory universe." (Wheeler, 1990)

We can find the relational view of information in Floridi's work (Floridi, 2003, 2008b, 2008a) in the form of Informational Structural Realism (ISR). This relational view in physics is formulated by Carlo Rovelli, (Rovelli, 2015) (Rovelli, 2018). Even Stephen Wolfram's "Project to Find the Fundamental Theory of Physics" (Wolfram, 2020) is based on the relational view, as well as the approach of (Chaitin, 2006) (Chaitin, 2018).
Barry Cooper (Cooper, 2015) addresses the issue of informational structures, citing (Floridi, 2011):

> "Structural objects work epistemologically like constraining affordances: they allow or invite certain epistemic constructs (they are affordances for inforgs like us, who elaborate them) and resist or impede some others (they are constraints for the inforgs), depending on the interaction with, and the nature of, the informational organisms that process them. They are exploitable by a theory, at a given LoA, as input of adequate queries to produce information (the model) as output." (p. 340)

and

> "all that physics – or science in general – tells us about, and hence all that we can know, on the basis of that physics, is structure." (p. 343)

This question of informational structures that are not "flat" is directly connected to models of computation as elaborated in the article "From the closed classical algorithmic universe to an open world of algorithmic constellations" (Burgin & Dodig-Crnkovic, 2012), indicating fundamental connections between information and computation as well as the importance of morphological (structural and material) aspects of both.

Matter and energy are higher-order concepts derived from interactions between observers and the universe, including mutual interactions among observers. This does not imply that reality is merely an agreed-upon fiction. The world exists independently, with human observers/agents discovering its structures and processes through interactions and self-structuring of information. The reality for a cognitive agent is built upon external information and the agent's information processing architecture, leading to diverse realities for different types of agents (Uexküll, 1957). Every subsystem in the universe interprets every other subsystem in a way dependent on their information processing capacities, (Friston et al., 2010) (Friston et al., 2010)(K. Friston & Kiebel, 2009)(Isomura et al., 2023).

All living beings have the ability to adapt to the environment and learn so as to increase their chances of survival. The life itself can be understood as a process of cognition (Maturana & Varela, 1980) (Stewart, 1996). Building on Maturana and Varela's understanding of cognition the author proposed an info-computational constructive framework (Dodig-Crnkovic, 2011b)(Dodig-Crnkovic, 2011a) to explain how increasingly complex structures develop as a result of information processing in nature. These information processes can be modeled as natural morphological computation, in the network of networks of physical processes where constant exchange (communication) of data (signals) among agents establishes new structures on physical, chemical, and biological levels. Instead of symbol manipulation, morphological computation is a result of physical object interactions. Natural computation (Rozenberg et al., 2012) can be used to explain the emergent phenomena by complexification of information through processes of signals exchange at different levels of organization or scale. The new framework is used as a tool for studying cognitive systems such as living organisms or artifactual cognitive agents on different levels of complexity.

For example, only recently we learned about the essential role of viruses and bacteria for the development of life on Earth, (Witzany, 2012). Agents exchanged among organisms can be bacteria, viruses, or pieces of DNA that can actively alter the organism that receives them, as they continue to act within an organism. All those processes of communication along with self-organization (autopoiesis) unfold in thermodynamically open living systems, and they need energy to persist, in the regime on the edge of chaos (Prigogine & Stengers, 1984) (Prigogine, 1980).

*Morphological info-computation in cognitive systems*

Over billions of years, nature has developed information processing networks in living organisms, enabling them to effectively cope with environmental complexity. The evolution from inanimate to animate matter, though largely unknown, suggests that self-assembly and self-organization of increasingly complex structures led to the first molecular machines and cells, eventually forming multicellular organisms. Communication processes and self-organization unfold in thermodynamically open systems, highlighting the significance of energy in maintaining these systems. Morphological computation, a result of physical object interactions, provides a framework for understanding the emergent phenomena in living systems (Dodig-Crnkovic, 2017)

*Evolution in the framework of info-computationalism/computational naturalism (ICON)*

Evolution in the framework of ICON is understood as the dynamic process of creating, transferring, and transforming informational structures across multiple scales and levels of organization in nature. This perspective views nature as a hierarchical network of informational agents engaged in computational processes, where information processing (computation) occurs. Morphological computation considers computation as the physical process of structural change in informational structures. Evolution is driven by these dynamic processes, where informational agents (such as cells or organisms) interact with their environment and with each other, creating new informational structures. Lower-level informational processes give rise to higher-level emergent phenomena through self-organization, with top-down control structures like the nervous system and the brain, regulating bodily functions.
The evolution of information is driven by multi-level interactions and information transfers between hierarchical levels, aligning with the principles of the extended evolutionary synthesis (EES) (Laland et al., 2015). (G. B. Müller, 2017) (Noble, 2002, 2012, 2016b, 2016a, 2020; Corning, Kauffman, et al. 2023) (Laland et al., 2015) (G. B. Müller, 2017) (Corning et al., 2023; Noble, 2002, 2012, 2016b, 2016a, 2020) (Torday et al., 2020; Torday & Miller, 2020).

The info-computationalism/computational naturalism/ICON framework provides a perspective on the evolution of complex systems by viewing them as hierarchical networks of informational agents engaged in morphological computation. Nature is organized into a dynamic hierarchy of levels, with informational processes (computation) occurring at each level, from elementary particles to biological systems. This hierarchical structure allows for the emergence of complex systems through interactions and information transfer across levels. The physical form (morphology) of informational agents and their interactions with the environment shape the computational processes underlying their behavior and intelligence. These interactions across genetic, epigenetic, behavioral, and cultural processes contribute to the emergence and evolution of complex systems.

By grounding cognition in morphological computation and embodied dynamics, the ICON framework provides a theoretical basis for rethinking and developing new approaches to artificial intelligence (AI) that better capture the embodied and situated nature of intelligence as observed in biological systems.

Traditional AI has typically focused on disembodied, symbolic reasoning systems, but the ICON framework suggests that truly intelligent systems should incorporate principles of embodied cognition, where the physical form and environmental interactions play a central role in shaping computational processes and cognitive behavior. Morphological computation could inspire the development of AI systems based on the computational capabilities of physical structures and their interactions with the environment, rather than relying solely on abstract algorithms and symbolic representations. This could lead to more robust, adaptive, and context-aware AI systems that better mimic the intelligence observed in natural systems.
Furthermore, the ICON emphasis on hierarchical organization, self-organization, and multi-level interactions suggests the idea of developing AI systems that can exhibit emergent behavior and adapt to complex environments through bottom-up and top-down interactions across multiple scales and levels of organization.

# The relationships between Magnani's Eco-cognitive computationalism and other views of cognition based on physical computation

The concepts of information, computation, and cognition are central to various branches of philosophy, cognitive science, and computer science. Lorenzo Magnani, Gordana Dodig-Crnkovic, Gualtiero

Piccinini, Oron Shagrir, Marcin Miłkowski, Nir Fresco and Michael Levin —each provide different perspectives on these topics.

*Lorenzo Magnani's Eco-Cognitive Computationalism* (Magnani, 2018) (Magnani, 2021) (Magnani, 2022b) (Arfini & Magnani, 2022) and Gordana Dodig Crnkovic's Info-Computationalism both study the integration of computation with cognitive and informational processes, but they focus on different aspects of this integration and have distinct foundational perspectives. Magnani is particularly interested in the concept of distributed cognition, which emphasizes that human cognition extends beyond the individual and into the environment and artifacts used. This approach views information as something not solely processed internally but also distributed across and manipulated by external media (Magnani, 2025).

*Eco-cognitive computationalism, emphasizes the role of computational tools* in reshaping human cognitive processes and environments. While it recognizes the importance of computational methods in extending and enhancing human cognitive capacities, *it is more focused on the interaction between humans and their cognitive tools in specific environments. Magnani's approach is particularly interested in how computational tools can domesticate complex, poorly understood, or "ignorant" entities, making them more accessible and manipulable by human cognition.*

*Info-Computationalism (ICON)*, (Dodig-Crnkovic, 2007) (Dodig-Crnkovic, 2008) ((Ehresmann, 2014), in contrast, views nature itself as computational, where information processing is fundamental to the physical, biological, and cognitive processes that constitute reality for a cognizing agent. This framework posits that all natural phenomena can be understood in terms of information processing and that knowledge generation is a result of computational processes in the natural world. The universe, from this perspective, is seen as a network of networks of computational processes where different levels of physical, biological, and cognitive phenomena interact through computational dynamics. ICON views information as a fundamental building block of reality for an agent, a perspective based on Informational Structural Realism (ISR) (Floridi, 2008a). In this interpretation, information, and computation are seen as essential for understanding natural and artificial processes, implying that computation is a broader phenomenon than what happens in digital computers. Cognition is intrinsically connected to the processing of information integrating this into the broader ecosystem, positioning cognition as an information-processing activity at all levels of biological and artificial systems (what Levin calls a "multiscale competency")(Levin, 2023). This approach introduces an integrative dimension to the study of cognition, with cognitive processes being more widespread in biology than previously thought. It supports a view similar to that of Magnani's distributed cognition, where cognitive processes are not confined to the brain but are distributed throughout the body and interact closely with the environment. The relation between these two theories can be seen in how they both extend the traditional view of cognition and incorporate computational thinking into the fabric of understanding and interacting with the world. However, while Eco-cognitive computationalism focuses on the enhancement of human cognition and the domestication of complexity through computational tools, Info-computationalism proposes a foundational theory where computation and information processing manifest in all natural processes.

Thus, while both frameworks elaborate the role of computation in understanding and interacting with the world, Eco-cognitive computationalism is applied and focused on the enhancement of human cognitive environments through specific computational interventions, whereas Info-computationalism provides a foundational perspective on nature as it appears to us, including psychological and social aspects. Both perspectives enrich our understanding of the computational nature of cognition and reality but from complementary angles.

*Gualtiero Piccinini's* approach offers another important dimension to consider regarding varieties of computationalism (Piccinini, 2012, 2020; Piccinini & Scarantino, 2011; Piccinini & Shagrir, 2014). Piccinini focuses on the philosophy of mind and the foundations of cognitive science, specifically the nature of computation in cognitive systems. He proposes a mechanistic view of computation, where a physical system is computational if it manipulates data through a set of functional transitions as specified by its architecture. Piccinini distinguishes between different kinds of computation based on how they are implemented physically and the kinds of processes they involve, such as digital versus analog. His theory is comprehensive in its definition and classification of computational processes, insisting on clear, functional, and mechanistic explanations to demarcate what counts as computation and what does not. Piccinini's emphasis on the physical basis of computation aligns well with Magnani's Eco-cognitive computationalism as both see computational tools as extensions and enhancements of human cognitive capabilities in tangible, real-world settings. Where Piccinini provides a framework for understanding how computational processes operate mechanistically within cognitive systems, Magnani looks at the outcomes of these processes—specifically, how they can domesticate "ignorant" entities to augment human cognitive capacities and modify cognitive environments.

*Oron Shagrir's* views on computation (Shagrir, 2010, 2022) further enrich the discourse, by addressing topics of physical computation, computational models, computational complexity, and hypercomputation (physically realizable computational models that can solve problems beyond the capabilities of Turing machines, including analog computation, quantum computation, and exploiting physical phenomena like chaos). Shagrir highlights the importance of mapping computational descriptions onto physical phenomena, asserting that a physical system is computational if there is a coherent way to map computational states onto the states of the physical system. His approach extends the understanding of computation beyond the strictly mechanistic framework of Piccinini, proposing a more inclusive approach that can accommodate different types of computations based on their functional realization in physical substrates. Shagrir extends the computational theory of mind by exploring how various types of computations can explain different cognitive abilities. He discusses information in the context of how it is represented and processed by the brain, viewing computation as a model for explaining how cognitive processes are implemented neurally. Shagrir provides a comprehensive and rigorous analysis of the physical foundations of computation, bridging the gap between theoretical computer science and the physical realizability of computational processes. Shagrir's way of approaching physical systems computationally complements Magnani's emphasis on using computational tools to extend human cognitive processes. In both frameworks, the physical instantiation of computation plays a central role.

*Marcin Miłkowski* (Miłkowski, 2013b) presents a rigorous analysis of explanation in cognitive science through his mechanistic theory of computation. He explores various computational explanations in cognitive science through four diverse case studies: classical computational simulation, connectionist modeling, computational neuroscience, and radical embodied robotics. Miłkowski outlines different computational processes, emphasizing that computation in cognitive science is broader than traditional symbolic representations. He argues for a mechanistic account of computation that includes digital, analog, and even quantum computation, highlighting the importance of information processing rather than mere symbol manipulation. He defends the view that computation in cognitive science is a physical process that underlies cognitive phenomena. Miłkowski points out the importance of a clear definition of computation that avoids pancomputationalism (the idea that everything computes). He advocates for the mechanistic framework as the most suitable for elucidating computational explanations, although he acknowledges the value of other explanatory frameworks. His approach to cognition is also mechanistic, viewing cognitive processes as computational operations defined by their causal structure and organization.

*Michael Levin* provides deep insights into cognition, particularly in non-neural biological systems (Baluška & Levin, 2016; Bongard & Levin, 2023; Fields & Levin, 2020; Levin, 2019, 2020, 2022, 2023; Levin & Pezzulo, 2016; Manicka & Levin, 2019, 2022; Mcmillen & Levin, 2024; Pezzulo & Levin, 2015). His work significantly expands the conventional understanding of cognition by studying how cells and non-neural organisms process information and exhibit behavior that suggests a form of basal intelligence. Drawing parallels between his work and the concepts of information, computation, and cognition, can provide a broadened perspective on these concepts. Levin's research primarily focuses on the bioelectric processes that underlie the decision-making and behavior in cells and tissues, independent of the presence of a nervous system. His studies have shown that cells and groups of cells can process information and make complex decisions, which are critical for morphogenesis, regeneration, and cancer suppression. This suggests a form of distributed cognition that occurs at cellular and tissue levels. In the context of information, Levin's work suggests that all living systems, not just neural systems, engage in processing environmental signals (information) to maintain homeostasis and to guide development and regeneration. This is akin to the information-centric view of ICON, where information is a fundamental component of all natural processes. Regarding computation, Levin demonstrates that non-neural and even single-cell organisms perform complex, goal-directed tasks by using bioelectrical signaling pathways to process inputs and drive outputs, effectively computing responses to their environment. Levin's research implies that biological computation can be seen in many forms and across various biological entities, supporting a broader, more inclusive definition of computation. Regarding cognition, Levin's work challenges the traditional brain-centered approaches. He proposes that cells and tissues exhibit a form of cognition, evidenced by their ability to process information, learn, make decisions, and remember states—behaviors typically reserved for organisms with brains. This aligns well with Karl Friston's Free Energy Principle, where organisms act to minimize surprise by forming predictive models of their environment. Levin's studies suggest that even single cells engage in predictive modeling and decision-making processes to minimize free energy, thus displaying a rudimentary form of cognition.

Levin's work shows that information processing and computation are basic life functions that occur at multiple levels of biological organization, thereby enriching our understanding of cognition in a broad biological context.

## Magnani on the interrelatedness of information, computation, and cognition

(Magnani, 2025) explores the interrelated concepts of information, computation, and cognition, highlighting their dynamic and evolving nature in the framework of Eco-cognitive computationalism. Magnani views these concepts from the perspective of ecological interactions between the brain, body, and environment, emphasizing their *coevolutionary aspects*. Information, computation, and cognition are conceptually deeply intertwined and influence each other's evolution.

*Magnani argues against viewing these concepts as static or isolated, suggesting instead that they dynamically interact and evolve within cultural, technological, and biological contexts.*

Eco-cognitive computationalism suggests that understanding computation should extend beyond traditional digital computation to include a variety of physical systems and processes. Computation is seen as active in various entities that have been transformed to encode and decode data, thus participating in cognitive processes. Semiosis (the process of signification or meaning-making) is fundamental to the development of information and cognition. Magnani integrates insights from processes of meaning-making from the catastrophe theory of René Thom - topology, morphogenesis, and structuralism (Aubin, 2020) in conjunction with Peircean semiotics to explain how biological and physical phenomena emerge as salient forms carrying information. Magnani discusses how changes in technology and cultural practices impact the meanings and functions of information, computation, and cognition. These changes facilitate the emergence of new forms of cognitive processes and information handling, reflecting a coevolutionary development between humans and their technological environments.

Magnani explores practical instances, such as how signs and tools externalize and distribute cognitive processes in the environment, a concept linked to extended cognition theories (Clark & Chalmers, 1998). He makes an important observation that *the meanings of information, computation, and cognition are not fixed but are subject to continual evolution and reinterpretation*. This emphasizes the need for an integrated approach that considers the *historical, ecological, and semiotic aspects* of these concepts to fully understand their roles and impacts in both natural and artificial systems. These concepts are not merely technical or theoretical but are deeply embedded in ecological, semiotic, and cognitive frameworks.

# Karl Friston's work related to information, computation, and cognition

Karl Friston's work, with his formulation of the Free Energy Principle (FEP), provides a perspective that relates to the concepts of information, computation, and cognition. Friston's theory is influential in neuroscience, psychology, and cognitive sciences, and it offers a unifying framework that can be related to the views of the other scholars mentioned earlier.

The Free Energy Principle suggests that all living systems (from single-cell organisms to human brains) aim to minimize their free energy. In this context, free energy is a mathematical concept borrowed from statistical physics but *represents an upper bound on the surprise or unpredictability of sensory data given a model of the world*. Minimizing free energy can be understood as *minimizing uncertainty or prediction error about the sensory inputs received from the world.*

In terms of information, Friston's approach can be seen as treating *cognition as fundamentally an information-processing problem*. The brain continuously infers and updates its internal models of the world to minimize the informational discrepancy (or "surprise") between its predictions and the actual sensory input. This aligns with views like those of (Dodig-Crnkovic, 2012a, 2014c, 2014a), who also see information as essential to understanding of cognitive processes.

Computation in Friston's theory is about the *neural mechanisms* that the brain employs to achieve this minimization of free energy. It involves complex computational strategies such as *predictive coding, where the brain uses hierarchical models to predict sensory inputs and adjust its beliefs based on prediction errors*.

The FEP effectively offers a theory of how cognition functions: it is the process by which the brain models the world to minimize the surprise associated with new sensory inputs. Cognition, therefore, is about reducing uncertainty through perpetual inference and learning. Predictive coding is a central aspect of Karl Friston's work, and a theory primarily concerned with how the brain processes *sensory* information. The key idea is that the brain is constantly generating and updating a mental model of the environment. This model is used to predict sensory input before it arrives. The brain then compares the actual sensory input with its predictions and adjusts its model based on the differences. The process is *hierarchical*. Higher levels of the neural network make predictions about what lower levels will perceive. Lower levels send back the prediction error (the difference between the expected sensory input and the actual sensory input) to higher levels. This feedback leads to the updating of the predictive model. At each level of this hierarchy, the goal is to minimize prediction error. *Minimizing prediction error can be equated to minimizing free energy*. Essentially, by minimizing the surprise or the prediction error, the brain is trying to improve its model of the world to better predict future sensory inputs.
Predictive coding has implications for learning and adaptation. As the brain continuously adjusts its model to minimize prediction error, it learns about the environment. This learning is the brain's way of becoming more efficient at predicting the world, which leads to better decision-making and behavior.
Neurobiologically, predictive coding is supported by the structure and function of the brain's cortex. Neurons in higher cortical layers are suggested to represent the predictions, while lower layers represent the prediction errors. These layers interact dynamically to update and refine predictions through synaptic weight adjustments, reflective of learning. Predictive coding has implications not only for neuroscience but also for psychology and cognitive sciences, explaining phenomena such as perception, motor control, and even psychiatric disorders. Predictive coding offers a comprehensive and computationally detailed account of how the brain might function as a prediction and error correction machine, within the Free Energy Principle.

Friston's framework thus provides a biological and theoretical foundation for discussing cognition as a process rooted in the principles of information theory and computational neuroscience. It bridges the gap between abstract computational theories and concrete biological mechanisms, offering a robust model that explains both the adaptiveness and the efficiency of cognitive processes in terms of energy minimization strategies.

Friston's work can be seen as both complementary to the interpretations of information, computation, and cognition discussed by Magnani, Dodig-Crnkovic, Piccinini, Shagrir, Miłkowski, and Levin. Each brings a unique perspective, but together, they enhance our understanding of the complex interplay between the brain, its computational processes, and the information it handles. Magnani focuses on how cognition extends beyond the brain to include interactions with the world. Piccinini emphasizes the mechanistic nature of computation in explaining cognitive functions. Shagrir elaborates on computational theories to account for various cognitive capabilities. Miłkowski provides a precise, mechanistic understanding of how computational processes underpin cognition. From the biological point of view, Levin expands the conventional understanding of cognition by studying how cells and non-neural organisms process information and exhibit behavior that suggests basal cognition and intelligence. Dodig-Crnkovic integrates information and computation as structure and process aspects into a naturalist framework. Their work collectively enriches the discourse of these foundational concepts, offering diverse yet complementary perspectives that advance our understanding of the mind and its processes.

## Future developments, cognitive computing, and AI

Eco-cognitive computationalism, proposed by Magnani, studies the role of the environment and external representations in shaping cognitive processes. It suggests that cognition is not only an internal process but involves interactions with the external world, including the manipulation of physical objects and representations. This perspective aligns with the embodied and situated cognition paradigm, which recognizes the importance of an agent's physical form and environmental interactions in shaping cognitive behavior. Similarly, the ICON (info-computationalism or computational naturalism) framework, discussed before, highlights the concept of morphological computation, where the physical structure and embodied interactions of agents shape their computational processes.

In the context of AI development, particularly for large language models (LLMs), eco-cognitive computationalism could inform the design of more grounded and context-aware systems. LLMs are trained on vast amounts of textual data, which means disconnected from the physical world of embodied experiences.

By incorporating principles from Eco-cognitive computationalism, LLMs could be designed to integrate external representations, such as images, diagrams, or physical objects, into their cognitive processes. This would involve multimodal learning, where the model learns to associate textual information with visual or other sensory inputs, potentially leading to more grounded and contextually relevant language understanding and generation.

Furthermore, Eco-cognitive computationalism emphasizes the role of external representations in cognitive offloading and distributed cognition. This could inspire the development of interactive AI systems that leverage external representations and tools to augment their cognitive capabilities. The Eco-cognitive perspective could inform the development of AI systems that can better adapt to and interact with dynamic environments, as it recognizes the importance of environmental factors in shaping cognitive processes. The principles of embodied and situated cognition, as well as the integration of external representations and environmental interactions, could potentially contribute to the development of more grounded, context-aware, and adaptive AI systems.

Similarly, the application of ICON framework to AI systems, with the principles of embodied and situated cognition, hierarchical organization, self-organization, information dynamics, morphological computation, and constructive approach offers insights and perspectives for rethinking the development of AI systems, including LLMs, to better capture the embodied and adaptive nature of intelligence observed in natural systems. Hierarchical Organization and Self-Organization could inform the development of architectures that incorporate hierarchical representations and self-organizing mechanisms, enabling the emergence of higher-level cognitive capabilities from lower-level interactions. Such architectures could potentially exhibit more flexible and adaptive behavior, akin to the self-organization observed in natural systems. The ICON framework adopts a constructivist approach, where meaning and semantics emerge from the interactions between agents and their environments, rather than being predefined or imposed externally. For LLMs, this could inspire the development of models that learn to construct meaning and semantics through interactions with multimodal data and environments, rather than relying solely on predefined symbolic representations. Such models could potentially exhibit more flexible and context-aware language understanding and generation capabilities. The concept of morphological computation could inform the design of LLMs that leverage physical or virtual embodiments to shape their computational processes. Natural computation is known for its energy efficiency and energy consumption and resource efficiency may be the decisive factors for the future of AI.

## Computationalism, its varieties, its critics and its raison d'être

The current article presents a new approach to cognition, framed in terms of information and computation within the broader context of computing nature, specifically focusing on morphological (physical, embodied) computation. It situates Lorenzo Magnani's Eco-cognitive computationalism within this wider discourse. Given the persistent critiques against computationalism since its inception, one might wonder how anyone could venture to build upon the "quicksand" of computationalism—particularly in its pancomputationalist form.

Computationalism in different forms has been criticized by so many, that it is impossible to address all those criticisms. Gualtiero Piccinini e.g. provides a systematic critique of limited, unlimited, and ontic Pancomputationalism (Piccinini, 2017) while the new book by Piccinini and Anderson provides a critique of Ontic Pancomputationalism (Anderson & Piccinini, 2024). Piccinini advocates his own mechanistic version of computationalism with the requirement of "robust mapping". Under this interpretation not even neurons compute as they do a variety of other life-sustaining tasks which do not directly contribute to cognitive computing as described in Neural Computation and the Computational Theory of Cognition by the same author, (Piccinini & Bahar, 2013).

Nir Fresco in Physical Computation and Cognitive Science (Fresco, 2014), and Marcin Miłkowski (Miłkowski, 2013)(Miłkowski, 2018), navigate carefully avoiding Scylla and Charybdis of pancomputational grounding, staying with the proposals that always leave space for *processes in nature that are not computational*.

Dominic Horsman, Viv Kendon and Susan Stepney propose a concrete schema to explicitly define conditions under which physical system computes in (Horsman et al., 2017). Similarly, the recent work done by physicists proposes "a category theory-based framework to redefine physical computing in light of advancements in quantum computing and non-standard computing systems" (Dehghani & Caterina, 2024) asking what physical systems can be considered computers and what constraints enable physical systems to compute.

The Church–Turing-Deutsch principle, posits computation as inherently physical, (Deutsch, 1985).

On the other hand, Giuseppe Longo compares (certain applications of) computationalism to scientism and mysticism, (Longo, 2020). He critiques symbolic, discrete computation but embraces geometry (and by extension, continuous computation) as crucial for understanding embodied cognition, (Longo, 2024). Longo suggests that cognition cannot be reduced to symbolic manipulation but instead involves continuous, analog processes that are central to how humans perceive and act in the world.

There are even authors defending new directions of computationalism, like Matthias Scheutz (Scheutz, 2002) who in an edited collection of essays shows how computationalism evolved from the simplistic idea that the mind is computable on the symbol manipulating Turing-Machine, to concurrent forms of embodied computation beyond the abstract Turing Machine model.

One recurrent theme is that the assertion "*all of nature continuously computes*" is a vacuous statement. Similar holds for the statement that "*all of nature is an informational structure*", a position known as Informational Structural Realism (ISR).

In the dialogue article (Dodig-Crnkovic & Miłkowski, 2023) I replied to Marcin Miłkowski's objections against computing nature / natural computationalism. Likewise, in the earlier dialogue article, Vincent Muller presented similar objections to pancomputationalism (Dodig-Crnkovic & Müller, 2011). The main stumbling block is that the assertion: "*nature is an info-computational phenomenon for an observer*" evokes fears of a "grey goo" vision where no distinctions can be made in that info-computational soup. However, I defend the position that natural computationalism is not trivial but a very productive approach where our present computing machinery is used to leverage discovery processes in nature organized in variety of levels of abstraction, with the help of a more general idea of natural computation as envisaged in (Rozenberg et al., 2012) and (Rozenberg & Kari, 2008).

In 2008 Luciano Floridi wrote ´A Defense of Informational Structural Realism´, (Floridi, 2008a) article defending ISR as a viable and coherent philosophical position that reconciles the strengths of both Epistemic Structural Realism (ESR) and Ontic Structural Realism (OSR), arguing that the world can be understood as a totality of informational objects interacting within a structural framework. Info-computationalism adds to this "pan-informational" structure its "pan-computational" dynamics.

In conclusion, the diversity of interpretations—ranging from mechanistic computationalism to natural computationalism—demonstrates the flexibility and relevance of this framework. The ongoing dialogue between critics and proponents shows that computationalism's raison d'être lies in its ability to provide new perspectives on cognition, nature, and the role of information. While critiques are numerous and necessary to refine the theory, the continued development of embodied, physical, and natural computational paradigms ensures that computationalism will remain a productive and evolving area of inquiry.